\documentclass[conference]{IEEEtran}
\usepackage[utf8]{inputenc}
\usepackage[T1]{fontenc}
\usepackage{times}
\usepackage{microtype}
\usepackage{amsmath,amssymb,bm}
\usepackage{graphicx}
\usepackage{booktabs}
\usepackage{multirow}
\usepackage{makecell}
\usepackage{array}
\usepackage{xcolor}
\usepackage{enumitem}
\usepackage{hyperref}
\usepackage{balance}
\usepackage{caption}
\usepackage{dblfloatfix}
\usepackage{cuted}
\hypersetup{colorlinks=true,linkcolor=blue,citecolor=blue,urlcolor=blue}
\title{Disentangling Visuo-Tactile Foresight:\\Oracle-Guided Interface Discovery for World Action Models}
\author{\IEEEauthorblockN{Zihang Yao\IEEEauthorrefmark{1}, Chaoyue Ding\IEEEauthorrefmark{2}, Yingying Yu\IEEEauthorrefmark{2}}
\IEEEauthorblockA{\IEEEauthorrefmark{1}Brigham Young University, Provo, Utah, USA}
\IEEEauthorblockA{\IEEEauthorrefmark{2}Beijing Academy of Science and Technology, Beijing, China}
\IEEEauthorblockA{\textit{Corresponding author: Yingying Yu}}}

\begin{document}
\maketitle

\begin{abstract}
Contact-rich manipulation remains challenging because successful control depends on physical interaction cues that are often weakly observable from vision alone. Recent tactile world action models jointly model future visual observations and tactile signals to guide action generation, but how such futures should be structured for effective use by the action expert remains underexplored. Directly studying this question with learned world action models is difficult because end-to-end behavior entangles physically invalid visual futures, unreliable predictions, inaccurate or cross-modally inconsistent tactile forecasts, and an unreadable future-to-action interface.

To make this interface independently studyable, we introduce Oracle Visuo-Tactile Foresight (OVTF), a controlled framework that supplies paired RGB and tactile futures from successful trajectories verified in simulation. By fixing the future provider, OVTF isolates the interface and asks a cleaner question: if the future is successful and physically executable, what representation allows the action expert to absorb its benefit? Within OVTF, we propose Asymmetric Phase-Local Future Memory (AFM), in which visual memory reads future vision, each tactile memory jointly attends to its own tactile stream and phase-aligned future vision, and cross-tactile access is blocked. We compare AFM with Modality-Isolated Future Memory (IFM), which removes visual-to-tactile access and processes each future modality independently.

Across seven tasks on the UniVTAC simulation benchmark, AFM achieves 32.0\% average success, compared with 23.7\% for IFM and 14.9\% for UniVTAC-ACT. This controlled comparison shows that selective phase-aligned visual--tactile routing provides a more actionable future-to-action bridge than complete modality isolation.
\end{abstract}
\begin{figure*}[t]
\centering
\includegraphics[width=0.99\textwidth]{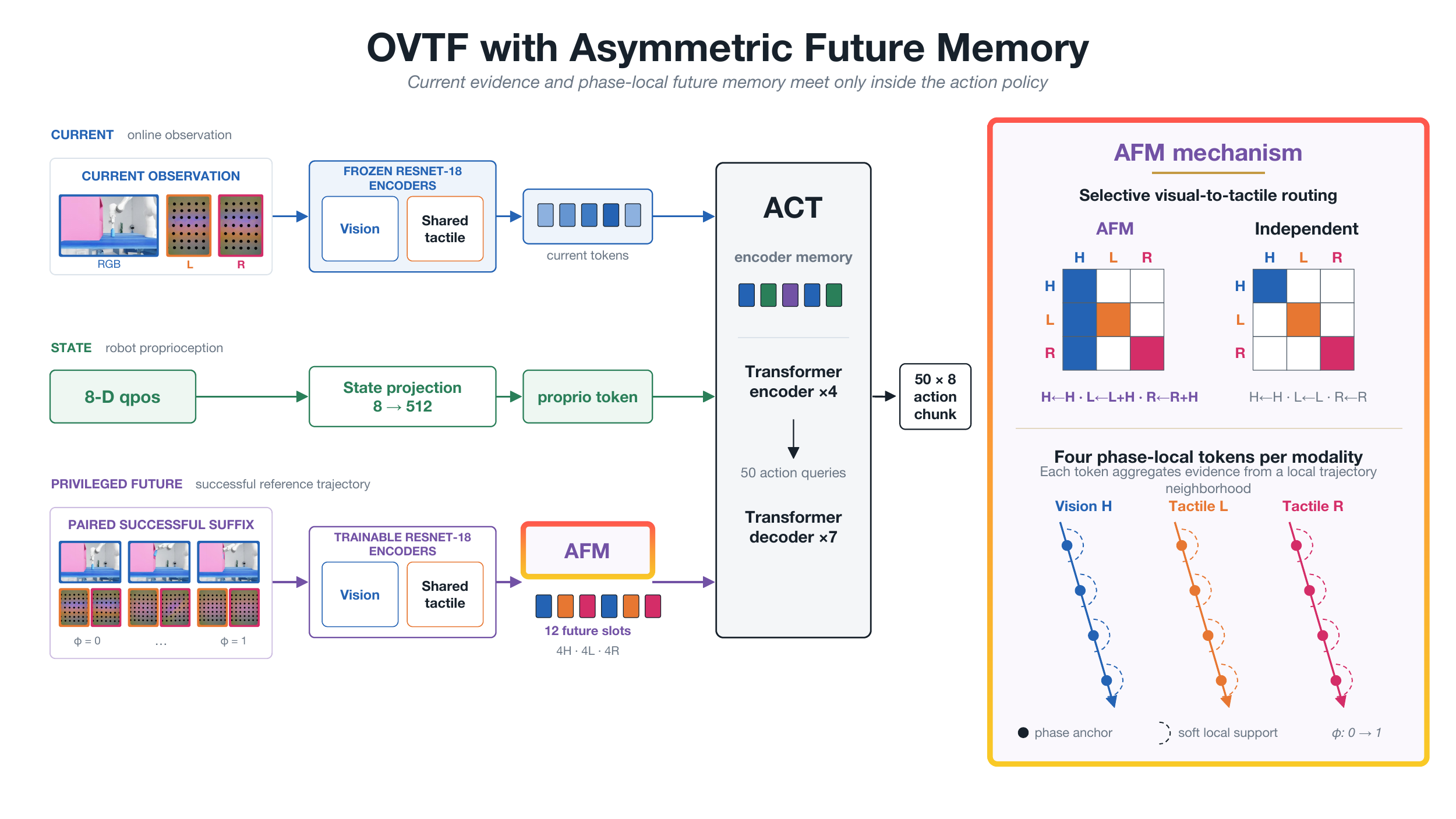}
\caption{OVTF with AFM. Current RGB and tactile observations are encoded by frozen UniVTAC-ACT* encoders, while 8-D qpos is projected into the shared ACT policy. Paired successful future RGB--tactile streams pass through trainable future encoders and AFM before entering the same policy. The inset compares AFM with the modality-isolated IFM route and visualizes phase-local aggregation. Although all memory tokens can interact again inside the ACT encoder, AFM explicitly performs asymmetric visual-to-tactile fusion at the actor-facing interface.}
\label{fig:method}
\end{figure*}

\section{Introduction}
Contact-rich manipulation remains challenging because successful control depends on physical interaction cues that are often weakly observable from vision alone. Recent tactile world action models extend predictive control by jointly modeling future visual observations and tactile signals to guide action generation in contact-rich tasks~\cite{dreamtac,tactilewam,vtwam,vtwm,dreamtacvla,omnivta,tacforesight}. However, the value of such foresight depends not only on whether the future is predicted accurately, but also on whether it is delivered through an interface that the action expert can effectively consume. The future-to-action interface is therefore a distinct bottleneck in visuo-tactile WAMs, yet it remains difficult to study in isolation.

Direct analysis with a learned WAM is challenging because four effects are entangled in the final task score. \textbf{(1) Physically invalid visual futures:} a visually convincing continuation may violate contact mechanics or describe a trajectory the robot cannot execute~\cite{vtwm,imagewam}. \textbf{(2) Insufficient trust in the future branch:} unstable or inaccurate predictions during training may teach the action branch to ignore future information, even when some predictions are useful~\cite{fastwam,imagewam}. \textbf{(3) Incorrect or mismatched tactile futures:} physically meaningful tactile prediction requires accurate contact timing, deformation, and local interaction dynamics, and jointly matching tactile evolution to a predicted video is especially difficult~\cite{vtwm,dreamtacvla,omnivta,tacforesight}. \textbf{(4) Poor future-to-action interface:} even a correct future may be encoded in a form that the action branch cannot efficiently read~\cite{oawam,kamwm,imagewam}. End-to-end WAM evaluation combines these effects, so it cannot reveal whether an interface itself is actionable.

Most existing WAMs optimize future prediction and action generation together. As long as future prediction remains imperfect, provider-side errors and consumer-side interface quality stay confounded, making it difficult to study how future information should be passed to the actor. We address this problem with Oracle Visuo-Tactile Foresight (OVTF). OVTF constructs a controlled future provider from successful demonstration trajectories whose RGB and tactile streams are paired, chronologically ordered, matched to the same initialization, and verified to succeed in simulation. By removing the learned predictor, OVTF makes the interface independently studyable and asks a cleaner question: \emph{if the future is successful and physically executable, what interface allows the action expert to absorb its benefit?}

Within OVTF, we propose Asymmetric Phase-Local Future Memory (AFM) and compare it with Modality-Isolated Future Memory (IFM). AFM allows visual memory to read future vision, allows each tactile memory to jointly attend to its own tactile stream and phase-aligned future vision, and blocks cross-tactile access. IFM retains the same future provider, phase anchors, memory budget, and optimization setting, but removes the visual-to-tactile reading edges. This comparison directly tests whether phase-aligned visual context makes future tactile information more actionable for action generation.

Across seven UniVTAC manipulation tasks, AFM with oracle futures reaches 32.0\% average success, compared with 23.7\% for IFM and 14.9\% for the UniVTAC-ACT* benchmark baseline~\cite{univtac}. The AFM--IFM gap shows that explicit asymmetric fusion at the future-to-action interface is more actionable than modality isolation, even though the resulting modality tokens can attend again inside the ACT encoder. AFM Zero-Train reaches 20.0\%, while the oracle-trained AFM retains 19.4\% when future inputs are disabled at evaluation; both remain above UniVTAC-ACT*, showing that the AFM-compatible policy path also performs strongly without future information.

\textbf{Contributions.} The core contributions of this paper are:
\begin{itemize}[leftmargin=1.2em]
    \item \textbf{Oracle Visuo-Tactile Foresight.} We introduce OVTF, a variable-disentangling framework that fixes the future provider with paired, simulation-verified successful RGB--tactile trajectories and asks a cleaner question: if the future is successful and physically executable, what interface allows the action expert to absorb its benefit?
    \item \textbf{An asymmetric future-memory interface.} We propose AFM and demonstrate that phase-aligned visual-to-tactile routing provides a more actionable pathway for future information consumption than IFM. Although the three modality memories can interact again inside the ACT encoder, explicitly performing asymmetric fusion at the actor-facing interface remains beneficial.
    \item \textbf{Strong performance without future inputs.} AFM Zero-Train and AFM Forced-Zero both outperform UniVTAC-ACT*, showing that the AFM-compatible policy path remains effective even when no future information is provided.
\end{itemize}

\section{Related Work}
\subsection{Vision-Language-Action and World Action Models}
Vision-language-action systems learn direct mappings from current observations and language instructions to robot actions. A complementary line of work uses video generation or predictive visual modeling to represent environment dynamics, synthesize plans, supervise action learning, or provide future-conditioned control~\cite{unipi,gr1,robodreamer,dreamitate,dreamzero,fastwam,oawam,kamwm,imagewam,visualforesight}. These approaches range from generating executable video plans to jointly predicting future observations and actions, or compressing predictive representations into actor-facing context. Our study differs by explicitly separating the future provider from the future-to-action interface: instead of asking whether a learned predictor and policy work jointly end-to-end, we ask what interface becomes most useful once the future is fixed. We use the ACT implementation and task checkpoints distributed with UniVTAC and denote this benchmark instantiation as UniVTAC-ACT* throughout this paper, where the asterisk distinguishes the UniVTAC benchmark policy from the original ACT formulation~\cite{univtac,act}.

\subsection{Visuo-Tactile Foresight and Tactile World Modeling}
Predictive tactile and visuo-tactile models use future touch, future visual--tactile dynamics, or contact-aware latent prediction to improve contact-rich manipulation~\cite{vitacformer,dreamtacvla,vtwm,omnivta,tacforesight,dreamtac,tactilewam,vtwam}. These works establish the value of tactile foresight, but typically evaluate the future provider, fusion mechanism, and action learner together. OVTF instead fixes paired visual--tactile future content and studies the actor-facing interface under controlled routing and future-availability conditions.

\section{Method}
\subsection{Oracle Future Source and Encoding}
At each policy call, the current observation contains task-dependent head and wrist RGB views, left and right tactile RGB images, and an 8-D joint-position vector $q_t$. Insert Tube and Lift Can use both head and wrist cameras; the remaining tasks use the head camera. All tasks use both tactile streams.

The oracle is not a contiguous prediction of $t+1{:}t+H$. It is sampled from a successful HDF demonstration trajectory paired with the current initialization. During training, for an action-window reference index $\tau_{\mathrm{ref}}$, we uniformly sample at most $N=100$ frames from $\tau_{\mathrm{ref}}$ (inclusive) to the final trajectory frame $T-1$. For sampled index $\tau_i$, the normalized remaining-horizon phase is
\begin{equation}
\phi_i = \frac{\tau_i-\tau_{\mathrm{ref}}}{\max(1,T-1-\tau_{\mathrm{ref}})},
\qquad \phi_i\in[0,1].
\label{eq:phase}
\end{equation}
During rollout, the policy-call index $p$ sets
\begin{equation}
\tau_{\mathrm{ref}}=\min\!\left(\left\lfloor p/2\right\rfloor,T-1\right),
\label{eq:rollout_ref}
\end{equation}
after which up to 100 remaining reference frames are sampled using the same phase definition. We do not infer a physical sampling frequency from this indexing rule.

The oracle provides only reference-trajectory visual images and left/right tactile RGB. It does not provide future actions, future qpos, object pose, simulator state, success labels, or other privileged metadata. It is therefore a privileged diagnostic source, not a learned future predictor.

Each future visual frame and tactile frame is encoded by a separate trainable ResNet-18 \cite{resnet}. These encoders are initialized by deep-copying the released UniVTAC-ACT* current vision/tactile encoder weights~\cite{univtac}. After spatial pooling and projection, the future provider yields three sequences,
\begin{equation}
H,L,R\in\mathbb{R}^{N\times512},
\end{equation}
not three pooled tokens. For Insert Tube and Lift Can, head and wrist images are encoded separately and fused by a two-layer MLP into a fixed 512-D $H$ stream.

Each sequence $X_m$, where $m\in\{H,L,R\}$, then receives a gated local temporal residual:
\begin{equation}
X'_m = X_m + \tanh(g_m)\,\mathrm{Conv}_{\mathrm{DW}}\!\left(\mathrm{LN}(X_m)\right),
\label{eq:temporal_filter}
\end{equation}
where $\mathrm{Conv}_{\mathrm{DW}}$ is a depthwise Conv1d with kernel size 5, padding 2, and 512 groups. The three gates $g_m$ are initialized to zero, suppressing the newly introduced future residual at initialization.

\subsection{Asymmetric Phase-Local Future Memory}
AFM contains 12 learned 512-D slots: four visual slots, four left-tactile slots, and four right-tactile slots. Each modality uses the fixed phase anchors
\begin{equation}
\mathcal{C}=\{0.125,0.375,0.625,0.875\}.
\end{equation}
For each allowed source sequence, one joint multi-head attention operation is performed. For attention head $h$, slot $k$, and valid source position $i$,
\begin{align}
\mathrm{score}^{h}_{k,i}
&= \frac{(q^h_k)^{\top}k^h_i}{\sqrt{64}}
-\frac{(\phi_i-c_k)^2}{2\times0.18^2}, \\
\alpha^{h}_{k,i}
&= \mathrm{softmax}_i\!\left(\mathrm{score}^{h}_{k,i}+\mathrm{mask}_i\right), \\
s_k
&= \mathrm{LN}\!\left(
W_o\,\mathrm{Concat}_h
\sum_i\alpha^{h}_{k,i}v^h_i
\right).
\label{eq:slot_attention}
\end{align}
The Gaussian term is a soft phase prior, not a hard temporal window. Content attention can still select among all valid source positions.

The AFM routing pattern is
\begin{equation}
H\leftarrow H,\qquad L\leftarrow(L,H),\qquad R\leftarrow(R,H),
\label{eq:apfm_route}
\end{equation}
with $L\leftrightarrow R$ access prohibited. Each left or right tactile slot performs a single joint attention operation over the concatenation of its own tactile sequence and the visual sequence; it does not first read tactile and then read vision. IFM uses
\begin{equation}
H\leftarrow H,\qquad L\leftarrow L,\qquad R\leftarrow R.
\label{eq:ind_route}
\end{equation}
The future-memory slots, phase anchors, model parameterization, future source, training data, and optimization budget remain fixed; the only routing change is removal of the $H\to L$ and $H\to R$ readable edges. Fig.~\ref{fig:method} visualizes both $3\times3$ routing matrices and the four Gaussian phase anchors.

\subsection{ACT Integration and Objective}
Current visual and tactile observations are processed by frozen UniVTAC-ACT* encoders, while qpos is processed by the frozen state projection, producing the current spatial memory. Each future slot is multiplied by a zero-initialized $\tanh$ gate. The implemented ACT encoder memory is organized as
\begin{equation}
[\,z_0+s_1,\; q,\; s_2,\ldots,s_{12},\; M_{\mathrm{current}}\,],
\label{eq:memory_order}
\end{equation}
where $z_0$ is the zero-latent base, $s_j$ is the $j$th gated slot, $q$ is the proprioception token, and $M_{\mathrm{current}}$ denotes the current spatial tokens. Action queries therefore read current observations and future slots through the same ACT encoder memory.

The ACT decoder uses 50 action queries and outputs a $50\times8$ normalized action chunk. During rollout, overlapping chunks are aggregated using weight $\exp(-0.01\times\mathrm{age})$. AFM disables the CVAE/action encoder during both training and inference; latent $z$ is always zero. Current encoders are frozen. Future encoders, temporal filters, the slot reader, ACT transformer, and action head are trainable.

The sole optimization objective is masked action $L_1$; slot decorrelation is detached and non-backpropagating:
\begin{equation}
\mathcal{L}_{\mathrm{action}}=
\mathrm{mean}\!\left(
|\hat a-a|\odot\mathbf{1}_{\mathrm{not\mbox{-}pad}}
\right),
\label{eq:loss}
\end{equation}

\newpage
\begin{strip}
\centering
\includegraphics[width=0.97\textwidth]{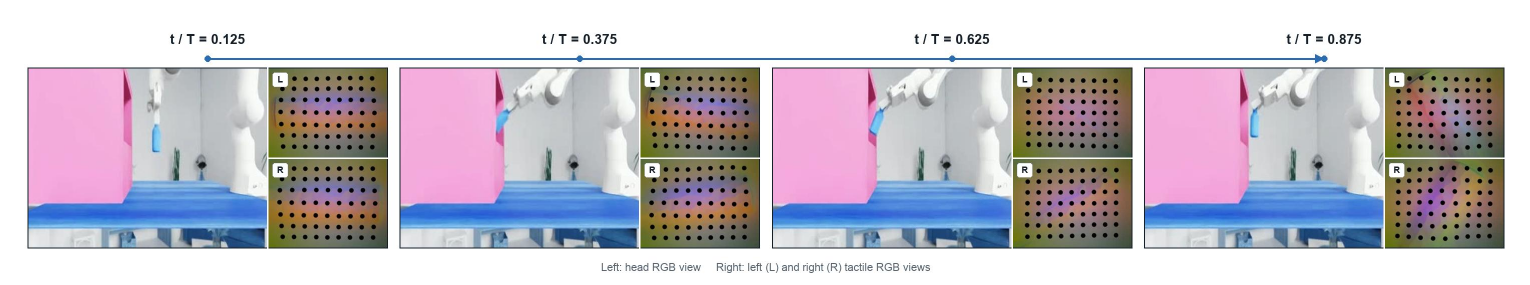}
\captionof{figure}{One real reference trajectory from the Put Shelf task. The four columns are sampled from the centers of four equal temporal segments of a successful trajectory and ordered from left to right. Each column shows the head RGB view on the left and the left/right tactile RGB views on the right. The visual sequence shows the blue bottle moving relative to the pink shelf, while tactile marker deformation reflects grasp and contact evolution. The figure is included only to illustrate task progression and the multimodal observation format.}
\label{fig:putshelfseq}
\vspace{2pt}
\captionof{table}{Results on the UniVTAC simulation benchmark \cite{univtac}. The highest score in each row is bold; ties for the highest score are all bolded. The second-highest distinct score is underlined; ties are all underlined. AFM Oracle is shown in the final column.}
\label{tab:main}
\small
\setlength{\tabcolsep}{4.6pt}
\begin{tabular}{lccccc}
\toprule
Task & UniVTAC-ACT* & AFM Zero-Train & IFM & AFM Forced-Zero & AFM Oracle \\
\midrule
Put Shelf & \underline{21/50, 42\%} & 14/50, 28\% & \underline{21/50, 42\%} & 17/50, 34\% & \textbf{27/50, 54\%} \\
Pull Out Key & 11/50, 22\% & \underline{15/50, 30\%} & 12/50, 24\% & 1/50, 2\% & \textbf{27/50, 54\%} \\
Lift Bottle & 1/50, 2\% & 2/50, 4\% & 2/50, 4\% & \underline{7/50, 14\%} & \textbf{8/50, 16\%} \\
Insert HDMI & \underline{11/50, 22\%} & 9/50, 18\% & \underline{11/50, 22\%} & 9/50, 18\% & \textbf{12/50, 24\%} \\
Insert Hole & 1/50, 2\% & 1/50, 2\% & \textbf{4/50, 8\%} & \underline{3/50, 6\%} & 2/50, 4\% \\
Insert Tube & 7/50, 14\% & 17/50, 34\% & \underline{20/50, 40\%} & \textbf{21/50, 42\%} & \textbf{21/50, 42\%} \\
Lift Can & 0/50, 0\% & 12/50, 24\% & \underline{13/50, 26\%} & 10/50, 20\% & \textbf{15/50, 30\%} \\
\midrule
Average & 52/350, 14.9\% & 70/350, 20.0\% & \underline{83/350, 23.7\%} & 68/350, 19.4\% & \textbf{112/350, 32.0\%} \\
\bottomrule
\end{tabular}
\end{strip}

\section{Experiments}

\subsection{Tasks and Splits}
We evaluate seven manipulation tasks from the UniVTAC simulation benchmark~\cite{univtac}. Every task uses both left and right tactile streams. Put Shelf uses the head camera, episodes 0--39 for training, episodes 40--49 as a development split, and episodes 50--99 for rollout evaluation. Pull Out Key, Lift Bottle, Insert HDMI, and Insert Hole use the head camera with episodes 0--49 for training and 50--99 for rollout evaluation. Insert Tube and Lift Can use both head and wrist cameras with episodes 0--49 for training and 50--99 for rollout evaluation.

\subsection{Fairness and Evaluation}
Within each task, all AFM-family arms use the same 50-row episode/reset mapping, outer evaluator, preprocessing, normalization, temporal aggregation, rollout horizon, early stopping, camera contract, and immutable vendor success predicate. The only permitted differences are the checkpoint and the minimum adapter required to express the designated future-information condition.

\subsection{Evaluation Conditions}
We evaluate four AFM-family conditions. \textbf{AFM Oracle} reads oracle futures during training and evaluation with asymmetric routing. \textbf{IFM} changes only the routing to Eq.~\eqref{eq:ind_route}. \textbf{AFM Forced-Zero} uses the Oracle-trained checkpoint but zeros all future-memory slots at evaluation. \textbf{AFM Zero-Train} never reads future HDF data and injects literal $12\times512$ zeros before the future modules throughout training and evaluation.

\subsection{Training and Implementation Details}
All AFM-family models are initialized from the corresponding task's released UniVTAC-ACT* checkpoint~\cite{univtac} and trained for 2000 optimizer steps on two NVIDIA RTX 4090 D GPUs. The per-GPU micro-batch is 1, gradient accumulation is 8, and the effective batch size is 16. We use AdamW with learning rate $10^{-5}$ and weight decay $10^{-4}$ for ordinary parameters; gate parameters use learning rate $10^{-3}$ and no weight decay. Cosine annealing ends at $10^{-6}$, and gradients are clipped at 1.0. Diagnostics are recorded every 100 steps, checkpoints are saved every 500 steps, and all reported AFM-family results use step 2000. Chunk size is 50, action dimension is 8, hidden dimension is 512, the transformer uses 8 heads, the ACT encoder has 4 layers, the decoder has 7 layers, and dropout is 0.1. Training uses FP32 without AMP. Future-encoding chunk size 8 is a memory optimization and does not change the model definition.

Experiments run on Ubuntu 20.04.6 with an Intel i9-14900K, 128 GB RAM, two NVIDIA RTX 4090 D 24 GB GPUs, driver 570.133.20, Python 3.10.12, PyTorch 2.7.0+cu128, torchvision 0.22.0+cu128, Isaac Sim 4.5.0, and Isaac Lab 0.41.3.

\subsection{Main Results}
Table~\ref{tab:main} reports the seven-task results for UniVTAC-ACT* and the four future-interface conditions. AFM Oracle achieves 112/350 = 32.0\% average success, compared with 52/350 = 14.9\% for UniVTAC-ACT*. These results show that successful future visuo-tactile signals provide important guidance for contact-rich action generation.

\subsection{AFM Family Ablation}
\begin{figure}[t]
\centering
\includegraphics[width=\columnwidth]{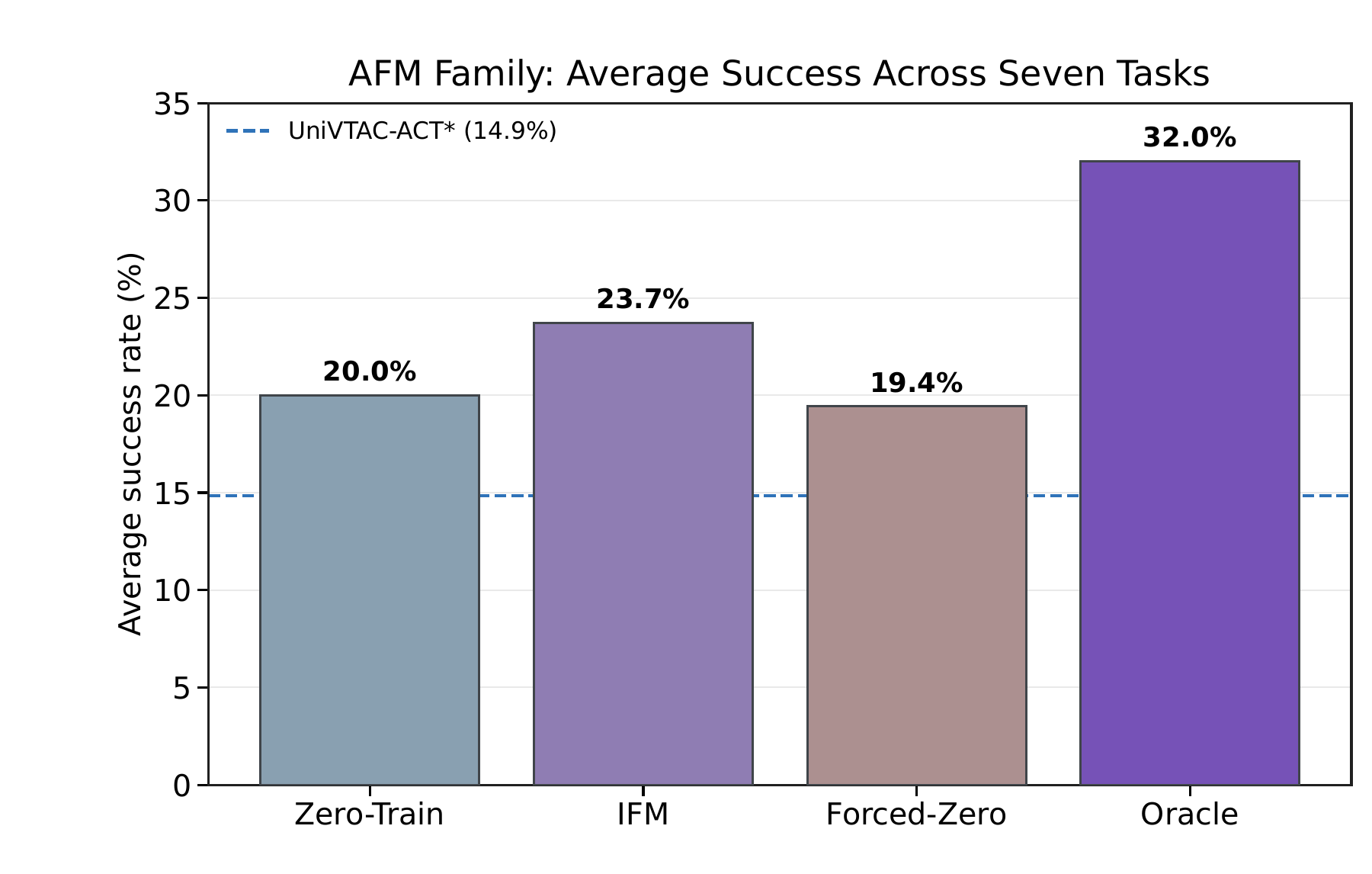}
\caption{Average success rate within the AFM family. The dashed line marks the 14.9\% UniVTAC-ACT* baseline.}
\label{fig:family}
\end{figure}
\begin{table}[t]
\centering
\caption{AFM-family average performance across 350 rollouts.}
\label{tab:family}
\small
\setlength{\tabcolsep}{4pt}
\begin{tabular}{lcccc}
\toprule
& Zero-Train & IFM & Forced-Zero & Oracle \\
\midrule
Successes & 70/350 & 83/350 & 68/350 & 112/350 \\
Average SR & 20.0\% & 23.7\% & 19.4\% & \textbf{32.0\%} \\
\bottomrule
\end{tabular}
\end{table}
Table~\ref{tab:family} and Fig.~\ref{fig:family} isolate the four AFM-family conditions. AFM Oracle reaches 32.0\%, whereas IFM reaches 23.7\% under the same oracle provider, future-memory budget, training data, and optimization budget. This 8.3-point gap shows that phase-aligned visual context makes tactile future memory more actionable than complete modality isolation. Importantly, all modality memories can still interact inside the subsequent ACT encoder; the improvement therefore shows that explicit asymmetric fusion at the interface contributes information organization that downstream self-attention does not recover from an isolated interface.

The same Oracle checkpoint falls to 19.4\% under Forced-Zero evaluation, establishing that AFM learns to use test-time future content rather than merely benefiting from additional modules. AFM Zero-Train reaches 20.0\%, and both evaluations without future information remain above the 14.9\% UniVTAC-ACT* baseline. These controls show that the AFM-compatible policy path also maintains strong performance when future information is absent, supporting the effectiveness of the framework itself.

\subsection{What OVTF Reveals}
OVTF separates three conclusions that a single end-to-end WAM score cannot expose. First, the Oracle--IFM comparison isolates future-interface routing and shows that asymmetric phase-aligned visual-to-tactile access is more actionable than modality-isolated memory. This remains true even though the modality tokens can subsequently re-attend inside ACT, demonstrating the value of explicit fusion at the interface itself. Second, the Oracle--Forced-Zero comparison shows that the oracle-trained policy genuinely uses test-time future content. Third, Zero-Train demonstrates that the AFM-compatible action path can operate without future HDF input and still outperform UniVTAC-ACT*.

The task-level results refine these conclusions. Put Shelf and Pull Out Key show the clearest behavioral value from future content. Insert Tube shows that a strong policy gain can persist without test-time future information, because Oracle and Forced-Zero tie at 42\%. Insert Hole is the only task where IFM is highest, but all conditions remain near the success-rate floor. Overall, the experiments establish both sides of the paper's claim: OVTF makes future-interface quality measurable, and AFM provides the strongest tested routing for converting paired visuo-tactile futures into action.

\section{Conclusion}
We introduced Oracle Visuo-Tactile Foresight, a variable-disentangling framework that fixes the future provider with paired, simulation-verified successful RGB and tactile trajectories. OVTF makes future-to-action interfaces independently studyable and asks which representation allows an action expert to benefit from a successful and physically executable future.

Using OVTF, we proposed Asymmetric Phase-Local Future Memory (AFM). AFM reaches 32.0\% average success, compared with 23.7\% for Modality-Isolated Future Memory (IFM) and 14.9\% for UniVTAC-ACT*. This controlled comparison establishes that selective phase-aligned visual-to-tactile routing forms a more actionable bridge than complete modality isolation. Although the modality memories can interact again inside the ACT encoder, explicitly organizing them through asymmetric fusion before policy integration remains beneficial. AFM Zero-Train reaches 20.0\%, and the Oracle-trained policy retains 19.4\% under Forced-Zero evaluation, showing that the AFM-compatible policy path also remains strong without future information. Together, these results show that visuo-tactile foresight becomes useful not only when the future is correct, but when it is organized through an interface designed for action generation.

\balance

\end{document}